\documentclass{article}
\usepackage[utf8]{inputenc}
\usepackage{authblk}
\usepackage{setspace}
\usepackage[margin=1.25in]{geometry}
\usepackage{graphicx}
\graphicspath{ {./figures/} }
\usepackage{subcaption}
\usepackage{amsmath}

\usepackage{hyperref}
\usepackage{amsmath,amsfonts}
\usepackage{graphicx}
\usepackage{booktabs}
\usepackage{multirow}
\usepackage{algorithm}
\usepackage{algpseudocode}
\usepackage{float}
\usepackage{arydshln}

\usepackage[style=nejm, 
citestyle=numeric-comp,
sorting=none]{biblatex}
\title{ProMUSE: Progressive Multi‑modal Uncertainty‑guided Staged Evidential Alzheimer Disease Classification}

\author[1$\dag$]{Long Doan}
\author[2$\dag$]{Branden Chen}
\author[3$\dag$]{Ethan Litton}
\author[4]{Huan Huang}
\author[5]{Jiajing Huang}
\author[6]{Yixin Xie}
\author[7]{Weihua Zhou}
\author[8]{Nandakumar Narayanan}
\author[9*]{Chen Zhao}

\affil[1,2,3,4,9]{Department of Computer Science, Kennesaw State University}
\affil[5]{Department of Data Science and Analytics, Kennesaw State University}
\affil[6]{Department of Information Technology, Kennesaw State University}
\affil[7]{Department of Applied Computing, Michigan Technological University}
\affil[8]{Department of Neurology, University of Iowa}
\affil[*]{Address correspondence to: czhao4@kennesaw.edu}
\affil[$\dag$]{These authors contributed equally to this work.}

\date{}

\begin{document}

\maketitle

\begin{abstract}
Alzheimer’s disease (AD) is a fatal disorder that destroys memory and cognitive skills in the elderly population. Most treatments for AD are effective in the early stage, leading to an increasing demand for early AD diagnosis. AD diagnosis increasingly relies on multi‑modal data such as clinical assessments, structural MRI (Magnetic Resonance Imaging), and PET (Positron Emission Tomography) imaging. However, MRI and PET acquisition remain costly and not universally accessible, making full‑modality inference impractical in real‑world clinical workflows. We propose \textbf{ProMUSE}, a \textbf{Pro}gressive \textbf{M}ulti‑modal \textbf{U}ncertainty‑guided \textbf{S}taged \textbf{E}vidential network that adaptively determines when additional modalities are necessary. which helps reduce the overall cost of data acquisition while maintaining the accuracy. ProMUSE first performs evidential classification using low‑cost clinical data and quantifies uncertainty via a Dirichlet‑based subjective logic model. When uncertainty exceeds a learned threshold, ProMUSE progressively incorporates MRI or PET features, fusing modality‑wise belief and uncertainty through Dempster–Shafer theory to obtain a calibrated multi‑modal prediction. This staged acquisition strategy enables accurate diagnosis while minimizing reliance on expensive imaging. Experiments on ADNI, AIBL, and OASIS across CN–AD, CN–MCI, and MCI–AD tasks demonstrate that ProMUSE achieves competitive or superior accuracy compared to full‑modality baselines while reducing MRI/PET usage by 50–90\%, yielding substantial cost savings. These results highlight ProMUSE as a practical, uncertainty‑aware, and resource‑efficient solution for real‑world AD screening.
\end{abstract}


\section{INTRODUCTION}
Alzheimer's disease (AD) is a neurodegenerative disorder that destroys memory and cognitive skills. The latest data from the CDC in 2022 showed that 11.3 percent of aging adults reported experiencing a decline in cognition and memory, with different rates across different age groups: 10.3\% of individuals aged 50 to 64 and 12.3 percent of those aged 65 and older \cite{matthews2019racial16}. A prediction from the Alzheimer’s Association indicates that 8 million Americans aged 65 and above will have Alzheimer’s disease by 2030 \cite{mitka2013pet17}. AD is one of the most common forms of dementia and is associated with the buildup of amyloid-$\beta$ plaques, tau tangles, and tau phosphorylation \cite{mahaman2022biomarkers18}. In general speaking, aging usually leads to Alzheimer's Disease, but there is no effective way to fully cure an Alzheimer's patient \cite{veitch2019understanding19}.

There are three main stages in the progression of AD: cognitively normal (CN), mild cognitive impairment (MCI), and AD. Among these, MCI represents a critical transitional stage, as individuals in this phase can benefit more from early interventions and therapeutic strategies compared to those with AD. However, MCI is challenging to distinguish from both CN and AD due to its subtle and heterogeneous characteristics. 

For AD classification, multimodal data are leveraged to capture complementary information. In this study, we utilize clinical data, magnetic resonance imaging (MRI), and positron emission tomography (PET) as three modalities. Clinical data are cost-effective and provide valuable information for assessing the risk of developing AD. MRI is effective in revealing structural brain changes, such as atrophy in the hippocampus and entorhinal cortex, but it lacks the ability to capture brain metabolism, which is a key biomarker of AD~\cite{dukart2013relationship6, chandra2019mri23}. To address this limitation, PET imaging is incorporated to characterize metabolic and functional brain changes that often precede cognitive decline~\cite{minoshima1997metabolic24}. Recent studies have shown that combining PET and MRI can improve AD diagnosis accuracy by approximately 3.0\% to 13.3\%~\cite{yang2022combining25, castellano2024automated26}. Therefore, we hypothesize that integrating clinical, MRI, and PET data can further enhance diagnostic performance.

Currently, most AD prediction models are primarily designed to maximize diagnostic accuracy, and they typically assume that all modalities are readily available at inference time. Under this paradigm, comprehensive medical tests, including MRI, PET scans, and even protein analyses, are implicitly required for every subject prior to prediction~\cite{song2021effective9, huang2021radiogenomics11, venugopalan2021multimodal29, zhang2019multimodal31}. While such approaches can achieve high predictive performance, they impose substantial financial and clinical burdens due to the high cost of acquiring all modalities. For instance, the cost of a PET scan starts at approximately \$3000 per visit~\cite{mitka2013pet17}, while an MRI scan in the United States typically ranges from \$558 to \$910. Moreover, from 1997 to 2010, the average cumulative cost of repeated diagnostic tests reached \$14,977~\cite{geldmacher2013prediagnosis32}. These observations highlight a key limitation of existing methods: by assuming full modality availability, they overlook the practical constraints of cost and accessibility in real-world clinical settings.

To address this challenge, we propose PROgressive Multi-modal
Uncertainty-guided Staged Evidential network (ProMUSE) that prioritizes inexpensive and widely accessible data sources before selectively incorporating costly modalities such as MRI and PET imaging. By leveraging uncertainty estimation, the proposed framework maintains diagnostic performance while significantly reducing reliance on high-cost resources. ProMUSE is evaluated on three independent datasets, namely Alzheimer's Disease Neuroimaging Initiative (ADNI) \cite{petersen2010adni46}, Australian Imaging, Biomarkers and Lifestyle Flagship Study of Aging (AIBL) \cite{ellis2009aiblDataset}, and Open Access Series of Imaging Studies (OASIS) \cite{marcus2010oasisDataset}, to assess both predictive accuracy and cross-dataset generalization. Experimental results demonstrate that the proposed ProMUSE achieves comparable diagnostic accuracy while reducing MRI/PET utilization by approximately 50\% to 90\%.

\section{MATERIALS AND METHODS}
\subsection{Related Work} 

\subsubsection{Multimodel integration methods for AD diagnosis}

In tasks with high complexity like AD diagnosis, a single omic is insufficient for performing accurate prediction. For that reason, multiple omics are utilized in these settings. To get the most from all modalities, a comprehensive fusion method is needed. Three famous types of multimodal fusion architectures are: early, intermediate, and late fusion \cite{zhou2024aidriven46}. Early fusion method is utilized due to its simplicity and superiority in capturing modality correlation. Intermediate fusion is superior to capture weak correlation between heterogeneous modalities, which is very useful in medical settings, where modalities' connection, even in different data formats, significantly increases the predictive powers. However, both early and intermediate are not as effective as late fusion in processing modality-specific features, which is prioritized in the ProMUSE framework, where clinical, MRI, and PET data provide different semantic information for the AD prediction. In late fusion, modality features are integrated after every modality is processed in different models. One of the common late fusion approaches is the voting mechanism, used in \cite{ayigit2022dementia47, gowda2022multimodal48}, to choose the final prediction by majority voting across processed uni-modality models. Another one is Ensemble Integration (EI), proposed by Yan Chak Li in \cite{li2022integratingEI}, forming uni-modality models into a global predictive model using an ensemble algorithm. These approaches help strengthen the predictive power of the model, but they can not evaluate the uncertainty of the model, which is the main feature for ProMUSE to work. For that reason, Dempster-Shafer Theory ~\cite{dempster200830-2} is adapted to ProMUSE to integrate single modality models for measuring the uncertainty degree of the model. 

\subsubsection{Staged method for multiview information fusion}

The ProMUSE's staged method is aimed at reducing the overall cost while maintaining the same accuracy. In terms of cost, most existing studies defined cost as the computational resources consumption instead of the monetary expense for data acquisition. One of the significant works is BranchyNet, which was proposed in \cite{teerapittayanon2016branchynet5-2}, introducing a new method to reduce the average computation by inserting an early-exit side branch when the decision confidence is high. Another representative work is BlockDrop, which was proposed in \cite{wu2018blockdrop12-2}, helping to reduce the computational cost by dynamically choosing residual blocks during the inference process. BranchyNet and BlockDrop effectively reduce the computational cost and time during inference, but the cost these models try to optimize is different from the cost in multi-omics medical settings. In medical scenarios, costs are the expenses coming from data acquisition and lab procedures. To switch the focus to real-world expenses, reinforcement learning (RL) is utilized for cost-aware prediction tasks. The balance between the predictive power and the data acquisition cost is the key purpose. \cite{contardo2016sequential6-2} makes a decision based on the data acquisition steps, where the model decides if extra resources need to be added based on the current amount of information. RL is integrated in \cite{an2022reinforcement7-2} to select lower-cost and informative features adaptive under resource constraints, which results in better prediction results. However, RL-based approaches make decisions based on a subset of features or individual features, which is not suitable for multi-omics medical settings, where every data acquisition results in a whole set of features. Additionally, the feature-level decision models fail to capture the fundamental difference between different modalities in terms of cost, turnaround time, and availability, making them differ from the real-world prediction process. ProMUSE solved all of these problems as it treats different modalities as fundamental decision units and prioritizes cheap, easily accessible resources, the same as the real-world diagnostics workflow.

\subsection{Methodology}

\subsubsection{Data Pre-Processing and Graph Modeling}
\label{sec:DataPre-Processing}

In this subsection, we preprocess all modalities to obtain trainable representations. Specifically, clinical data are transformed into feature vectors, while MRI and PET data are represented as graph-structured data. A graph is defined as $\mathbf{G = [E, X]}$, where $\mathbf{E}$ is the adjacency matrix of the connectivity of nodes and $\mathbf{X}$ is the features contained in each node. For MRI, each image is processed using FreeSurfer \cite{fischl2012freesurfer} to obtain brain masks with semantic segmentation. Based on these segmented regions, radiomic features as $\mathbf{X}_{mri}$ are extracted using pyradiomics \cite{zhao2021lung}. Each region corresponds to a graph node. The edges of the graph, as $\mathbf{E}$, are constructed based on the spatial proximity between nodes. For PET, the graph structure shares the same node coordinates as the MRI-derived graph to ensure anatomical consistency. The key difference lies in the node features as $\mathbf{X}_{pet}$, which are derived from the PET Unified Pipeline (PUP), including the total number of voxels ($N_{\text{vox}}$) and the Standardized Uptake Value Ratio (SUVR). 

In this study, clinical data is processed by Multilayer perceptron (MLP). Our MLP model has a total of three learnable layers comprised of 4 linear layer. The set of hidden units we use are 128, 256, 512 and 1024. Graph Neural Networks (GNNs) are deep learning architectures specifically designed to learn from graph-structured data by leveraging both node attributes and topological relationships. In this study, MRI and PET modalities are represented as graphs and processed using an identical GNN architecture.

The proposed model consists of three learnable layers: two GraphSAGE layers \cite{hamilton2017inductivegraphsage}, followed by a fully connected linear layer. To evaluate the impact of model capacity, the hidden dimensionality (h) is selected from the set ({128, 256, 512, 1024}), and all configurations are independently trained and evaluated. For a hidden dimension of (h), the first GraphSAGE layer maps the input node features from dimension (d) to (h), while the second GraphSAGE layer further transforms the node representations within the same (h)-dimensional latent space. The resulting graph embedding is then passed through a linear layer that reduces the feature dimension from (h) to (h/2). Finally, a classification layer produces two output logits corresponding to the target classes.

For example, when (h = 128), the network follows the transformation pipeline $(d \rightarrow 128 \rightarrow 128 \rightarrow 64 \rightarrow 2)$, where (d) denotes the input feature dimension.

\subsubsection{Uncertainty Quantification}
\label{sec:Uncertainty}

We propose a ProMUSE that estimates uncertainty across multiple modalities based on subjective logic, where uncertainty is explicitly modeled as a function of the available evidence. Subjective logic~\cite{josang201618-2} provides a principled foundation for uncertainty modeling through the Dirichlet distribution~\cite{cover199929-2}, enabling the representation of both belief masses and uncertainty. In our model, each class is associated with a certain amount of evidence, which is transformed into class-wise belief (confidence) via a Dirichlet-based formulation. The remaining unassigned probability mass is then interpreted as uncertainty, reflecting the amount of missing evidence.

In our staged method, output logits are extracted from clinical data, MRI, and PET modalities using GNNs as explained in Section \ref{sec:DataPre-Processing}. Each modality produces a logit vector that quantifies class-specific support. We adopt the Softplus activation function, which ensures non-negative outputs and allows unbounded growth, making it suitable for evidence modeling. The Softplus function is defined as $\mathrm{Softplus}(x) = \ln(1 + e^{x})$. For modality \(m\) and \(T\) target classes, the activation function converts the logit vector to corresponding evidence vector used for Dirichlet parameterization, denoted as $\mathbf{e}^{(m)} = [e^{(m)}_1, e^{(m)}_2, \ldots, e^{(m)}_t ]$. The corresponding Dirichlet parameters are defined in Eq. \eqref{eq:dir}: 

\begin{equation}
\begin{aligned}
\mathbf{d}^{(m)} 
&= [d^{(m)}_1, d^{(m)}_2, \ldots, d^{(m)}_t] \\
&= \mathbf{e}^{(m)} + 1\\
&= [e^{(m)}_1 + 1, e^{(m)}_2 + 1, \ldots, e^{(m)}_t + 1]
\end{aligned}
\label{eq:dir}
\end{equation}

In the next step, we compute belief mass ${B}^{(m)} = [B^{(m)}_1 ,B^{(m)}_2, \ldots, B^{(m)}_t]$ and uncertainty degree $U^{(m)}$ with the constraints that $U^{(m)} + \sum_t B^{(m)}_t = 1$, s.t. $U^{(m)} > 0$ and $B^{(m)}_t > 0$. Then, the sum of all Dirichlet parameters, $S^{(m)} = \sum_t d^{(m)}_t$, the belief mass is then derived as $B^{(m)}_t = \frac{e^{(m)}_t}{S^{(m)}}$. According to $S^{(m)}$, the uncertainty is denoted as $U^{(m)} = \frac{T}{S^{(m)}}$, where $T$ is the number of classes. With each modality, a vector of evidence is given. With informative and high-quality datasets, adding modalities can increase the amount of useful evidence, resulting in lower uncertainties.

\subsubsection{Dempster-Shafer theory for multi-modality uncertainty measurement}
\label{sec:Dempster}

The Dirichlet-based uncertainty estimation framework is inherently limited to single-modality inputs, which restricts its applicability to complex AD classification tasks. In these scenarios, relying on a single modality is often insufficient to capture the heterogeneous and complementary biomarkers of the disease. Specifically, the clinical data provides the risks of having AD, while MRI provides static brain visualization and PET provides brain metabolism. Any of these biomedical features significantly contributes to the predictive power of the model. To address this limitation, we further introduce the Dempster-Shafer Theory (DST)~\cite{dempster200830-2} to fuse evidence from multiple modalities and improve both prediction accuracy and uncertainty estimation.

Assume two sets of mass functions derived from two modalities, denoted as $M^1 = [B^1_1, B^1_2, \ldots, B^1_t, U^1]$ and $M^2 = [B^2_1, B^2_2, \ldots, B^2_t, U^2]$, where \(t\) represents the number of classes, \(B^m_k\) denotes the belief mass assigned to class \(k\) from modality \(m\), and \(U^m\) denotes the associated uncertainty mass. The fused mass function is defined as $M^{12} = [B^{12}_1, B^{12}_2, \ldots, B^{12}_t, U^{12}] = M^1 \oplus M^2$, where the combination follows the DST rule. The fused belief mass for class \(t\) is computed as shown in Eq. \ref{eq:6}.

\begin{equation}
\label{eq:6}
B^{12}_t = \frac{1}{1 - C} \left( B^1_t B^2_t + B^2_t U^1 + B^1_t U^2 \right),
\end{equation}

and the fused uncertainty is given by:

\begin{equation}
\label{eq:uncertainty_multi}
U^{12} = \frac{1}{1 - C} U^1 U^2,
\end{equation}
where \(C = \sum_{i \neq j} B^1_i B^2_j\) is the conflict coefficient, measuring the degree of disagreement between the two modalities.

The term \(\frac{1}{1 - C}\) serves as a normalization factor to ensure that the fused belief masses and uncertainty sum to one, thereby preserving probabilistic consistency. In addition, the cross terms \(B^2_t U^1\) and \(B^1_t U^2\) redistribute uncertainty into class-specific belief, enabling more robust evidence integration across modalities.

A high conflict value \(C\) indicates substantial disagreement between modalities, resulting in increased uncertainty and reduced predictive confidence. Conversely, a low conflict value implies consistent evidence across modalities, leading to reduced uncertainty and more confident predictions. This theory is a strong stepping stone for a cost-efficient framework in Alzheimer's disease. By combining the belief mass and uncertainty degree, more details can be captured, leading to better results in this task.

\subsubsection{Staged AD classification using multi-modality with uncertainty quantification}
\label{sec: full-frame}

\begin{figure}[H]
    \centering
    \includegraphics[width=0.5\textwidth]{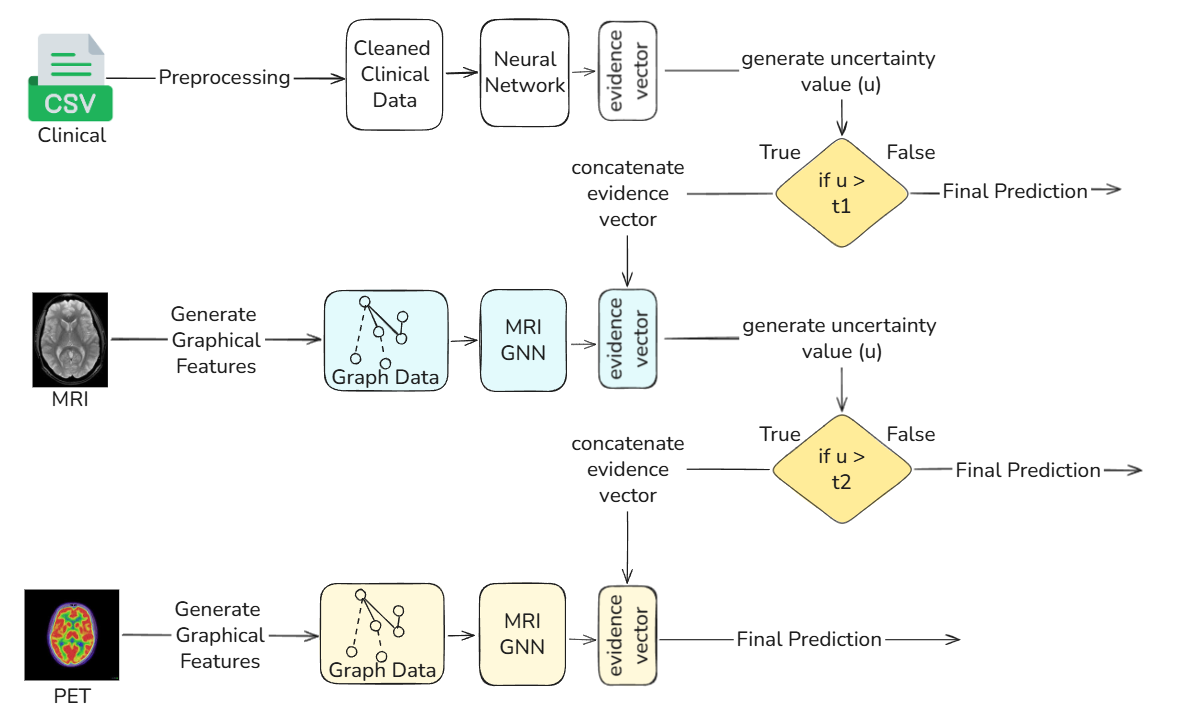}
    \caption{Multi-model uncertainty framework for prediction}
    \label{fig:uncertainty_framework}
\end{figure}

As shown in Fig. \ref{fig:uncertainty_framework}, ProMUSE performs the analysis and prediction using clinical, MRI, and PET. The first task is to collect raw data, such as clinical, MRI, and PET data. After that, raw data are processed to generate modality-specific uncertainty values. After generating the uncertainty value, if the uncertainty value is above the assigned threshold, another modality is added to perform another analysis and prediction. This cycle runs until all modalities are used or all uncertainty values are below the assigned threshold. With this framework, the accuracy of the model is well-maintained with fewer resources required. 

For this particular task, clinical data is the first modality to be analyzed. Among all 3 modalities we have mentioned, clinical data is the cheapest and most available modality, making it the most reasonable to start the framework. The raw clinical data is preprocessed into cleaned clinical data, which is analyzed by a neural network to output the evidence vector. The evidence vector is used to generate the uncertainty value. If the uncertainty value is below the assigned threshold (\(\tau_1\)), ProMUSE will perform the final prediction. Otherwise, if the assigned threshold (\(\tau_1\)) is exceeded by the uncertainty value, another modality, MRI, is added to perform further analysis. MRI addition forces the fusion between MRI and clinical data. MRI is the second choice, as its price is lower compared to PET. MRI is converted to graph data, which is analyzed by GNN to output the evidence vector. The evidence vector is used to generate the new uncertainty value. The detailed process is explained in Section \ref{sec:DataPre-Processing} and Section \ref{sec:Uncertainty}. The new uncertainty value is compared with the second assigned threshold (\(\tau_2\)). If the uncertainty value is higher, PET is added as the final modality to perform the final analysis; otherwise, the analysis stops, and the framework arrives at the final prediction. For the last step, the PET uncertainty value is calculated in the same way as MRI. In the fused step, clinical, MRI, and PET are fused together using the Dempster-Shafer method as explained in Section \ref{sec:Dempster}. As there is no more modality to add, this step always leads to the final prediction.

\subsubsection{Threshold Selection}

In the framework proposed in Section~\ref{sec: full-frame}, the threshold parameter $\tau$ plays a critical role in determining the model’s decision behavior. Notably, the optimal value of $\tau$ may vary across different datasets and tasks, depending on the underlying data distribution and classification difficulty. To address this issue, we design a data-driven procedure to select an optimal threshold $\tau$, as summarized in Algorithm~1. The proposed algorithm is specifically tailored to the characteristics of the dataset and task under consideration, enabling adaptive threshold selection that improves overall model performance and stability.

\begin{algorithm}[H]
\caption{Optimal Threshold Selection} \label{alg:threshold}
\begin{algorithmic}[1]

\State \textbf{Input:} \(u^c = [u^c_1, u^c_2, \ldots, u^c_n]\),
\(u^{cm} = [u^{cm}_1, u^{cm}_2, \ldots, u^{cm}_n]\),
correct\_labels

\State \textbf{Output:} optimal\_tau1, optimal\_tau2
\State range\_tau1 \(= linspace[min(u^c), max(u^c), 100]\)
\State range\_tau2 \(= linspace[min(u^{cm}), max(u^{cm}), 100]\)

\State optimal\_tau1 \(= 0\), optimal\_tau2 \(= 0\), best\_accuracy \(= 0\)

\State num\_patients \(=\) length of correct\_labels

\For{tau1 in range\_tau1}
\For{tau2 in range\_tau2}

\State predict\_labels \(= [\ ]\)

\For{p from 1 to num\_patients}

\If{\(u^c_p \leq \tau_1\)}
\State{
predict\_labels += prediction (clinical)
}

\ElsIf{\(u^{cm}_p \leq \tau_2\)}
\State{
predict\_labels += prediction (clinical + MRI)
}

\Else
\State{
predict\_labels += prediction (clinical + MRI + PET)
}

\EndIf
\EndFor

\State accuracy \(=\)
\(
\frac{
\text{correct\_ans(predict\_labels, correct\_labels)}
}{
\text{num\_patients}
}
\)

\If{accuracy \(>\) best\_accuracy}
\State optimal\_tau1 \(=\) tau1
\State optimal\_tau2 \(=\) tau2
\State best\_accuracy \(=\) accuracy
\EndIf

\EndFor
\EndFor

\end{algorithmic}
\end{algorithm}

\subsubsection{Loss Function}

The proposed model adopts a hybrid loss function consisting of two main components: a cross-entropy term for classification and a Kullback-Leibler (KL) divergence regularization term for uncertainty calibration. The cross-entropy loss is defined in Eq.~\ref{eq:ce_loss}.

\begin{equation}
L_{ce} = -\sum_{t=1}^{T} g_i^t \log(p_i^t)
\label{eq:ce_loss}
\end{equation}

\noindent where $p_i^t$ denotes the predicted probability of sample $i$ belonging to class $t$, and $g_i^t$ is the corresponding one-hot ground-truth label. While this formulation effectively supervises class prediction, it does not explicitly penalize or regulate model uncertainty. To address this limitation, we incorporate an evidential learning objective based on the Dirichlet distribution. The Dirichlet-based cross-entropy loss is defined as in Eq. \ref{eq:9}.

\begin{equation}
\label{eq:9}
L_{dce}(d_i) = \sum_{t=1}^{T} g_i^t \left( \psi(S_i) - \psi(d_i^t) \right),
\end{equation}

\noindent where $\psi(\cdot)$ denotes the digamma function, $d_i^t$ is the Dirichlet parameter (evidence) for class $t$, and $S_i = \sum_{t=1}^{T} d_i^t$ represents the total evidence. This objective encourages the model to assign stronger evidence to the correct class while implicitly modeling uncertainty through the Dirichlet concentration parameters.

In addition, a KL divergence regularization term is introduced to prevent the model from arbitrarily increasing evidence for all classes. The KL divergence between the predicted Dirichlet distribution and a non-informative prior is given by:
\begin{equation}
\begin{aligned}
KL\big[\mathrm{Dir}(\hat{d}) \,\|\, \mathrm{Dir}(1)\big]
&= \log \left(
\frac{
\Gamma \left(\sum_{t=1}^{T} \hat{d}_i^t \right)
}{
\Gamma(T)\prod_{t=1}^{T}\Gamma(\hat{d}_i^t)
}
\right) \\
&\quad + \sum_{t=1}^{T} (\hat{d}_i^t - 1)
\left[\psi(\hat{d}_i^t) - \psi(S_i)\right],
\end{aligned}
\end{equation}
where $S_i = \sum_{t=1}^{T} d_i^t$, and $\hat{d}_i = y_i + (1 - y_i) \odot d_i$ is a transformed Dirichlet parameter that assigns a non-informative state to the ground-truth class, thereby preventing penalization of correct predictions.

The final loss function for sample $i$ is defined as:
\begin{equation}
\label{eq:11}
l(d_i) = L_{dce}(d_i) + I_r \cdot KL\big[\mathrm{Dir}(\hat{d}) \,\|\, \mathrm{Dir}(1)\big],
\end{equation}
where $I_r$ is an annealing coefficient that controls the influence of the KL regularization term. Specifically, $I_r$ is initialized close to zero and gradually increases to one during training, allowing the model to first learn discriminative evidence before enforcing uncertainty regularization.

Finally, the overall training objective aggregates the losses from both individual modalities and the fused representation:
\begin{equation}
\label{eq:12}
L_{\mathrm{overall}} = \sum_{i=1}^{N}
\left[
L(d_i) + \sum_{m=1}^{M} L(d_i^m)
\right],
\end{equation}
where $d_i^m$ denotes the Dirichlet evidence vector for sample $i$ under modality $m$. This formulation ensures that both modality-specific and fused representations are optimized under a unified evidential learning framework.

\section{RESULTS AND DISCUSSION}
\subsection{Datasets and Experimental Settings}
We used three famous datasets to evaluate the effectiveness of our framework. These three datasets have different clinical data formats. The results for the framework performance are very positive. The datasets that we used for experiments are:

\begin{itemize}
\item \textbf{ADNI} \cite{petersen2010adni46}: This dataset contains a total of 432 patients with 1860 visits, including CN, MCI, and AD. Every visit has a fully detailed MRI, PET, and clinical data. This is a public-private dataset, launched in 2004, to help identify biomarkers for early detection of Alzheimer's Disease. 
\item \textbf{OASIS}\cite{marcus2010oasisDataset}: This dataset contains a total of 927 patients with 1595 visits, including CN, MCI, and AD.  Every visit has a fully detailed MRI, PET, and clinical data.
\item \textbf{AIBL}\cite{ellis2009aiblDataset}: This dataset contains a total of 582 patients with 1085 visits, including CN, MCI, and AD. Every visit has a fully detailed MRI, PET, and clinical data.
\end{itemize}

To fully capture all the details in the brain that cause AD, we have used 3 modalities for different kinds of features, including clinical variables, structural MRI, and PET features.

\subsection{Tasks}
For effective treatment and recovery, MCI is the best stage to do it.
Early intervention can significantly reduce the treatment cost for AD.
For that reason, the task to classify between healthy people and people with MCI and between MCI and full-blown Alzheimer's disease becomes very crucial to give out timely treatment. However, MCI is very difficult to pinpoint as it has many overlapping features with CN and AD. According to the above reasons, our experiments carry out three tasks to give more comprehensive results for the diagnosis process:
\begin{itemize}
\item \textbf{} CN vs AD
\item \textbf{} CN vs MCI
\item \textbf{} MCI vs AD
\end{itemize}

\subsection{Experiments Set-up}

To begin with, The word "CLI" under Modalities columns in Table \ref{tab:threshold_results} is the shortcut of clinical data Table \ref{tab:threshold_results}, the framework in Fig. \ref{fig:uncertainty_framework} is used to evaluate ADNI, AIBL, and OASIS with 3 different binary tasks (CN vs AD, CN vs MCI, MCI vs AD). For example, with CN vs AD in ADNI, the data is examined with only clinical data. 50\% of the samples' uncertainty values are below the first assigned threshold (0.495). The remaining 50\% is analyzed using both clinical and MRI data, resulting in 26.27\% of samples with uncertainty values below the second assigned threshold (0.015). All assigned threshold are found using Algorithm \ref{alg:threshold}. The accuracy, F1, and AUC are found using the subset of the dataset. For example, for CN vs AD in ADNI, 50\% of the dataset prediction using only clinical data, 26.27\% of the dataset prediction using clinical data and MRI, and 23.73\% of the dataset prediction using clinical data, MRI, and PET are combined to the final prediction of the whole dataset. The accuracy, F1, and AUC columns are calculated on this new combined prediction. The process is repeated for every combination of task and dataset. 

In Table \ref{tab:all_tasks_results_adni}, \ref{tab:all_tasks_results_AIBL}, \ref{tab:all_tasks_results_oasis}, clinical data is processed the same as ProMUSE, while MRI and PET graphs are flattened to a 1D data vector. For example, a graph with 84 nodes and 85 features in each node is converted to one row with 84*85 columns. Each dataset is split into 70\% train, 10\% validation, 20\% test. Each combination of dataset and task is trained and tested on 10 different models (KNN \cite{fix1950discriminatory34-2}, SVM \cite{vapnik199535-2}, LR \cite{tibshirani199636-2}, RF \cite{ho199537-2}, NN \cite{mcculloch194338-2}, GRridge \cite{van201639-2}, BPLSDA \cite{singh201940-2}, BSPLSDA \cite{singh201940-2}, MOGONET \cite{mogonet202122-2}, TMC \cite{han2022trusted10-2}, CF \cite{hong202041-2}, GMU \cite{arevalo201742-2}, MMDynamic \cite{han202243-2}, MLCLNet \cite{zheng202344-2}, DCP \cite{liu202145-2}, CLCLSA \cite{clclsa202423-2}). ProMUSE results are displayed in the last row of each dataset. Table \ref{tab:all_tasks_results_adni},\ref{tab:all_tasks_results_AIBL},\ref{tab:all_tasks_results_oasis} use Accuracy, F1-score, and area under the ROC curve (AUC) to reflect better on predictive power, robustness to class imbalance, and generalization ability. The best values in each column are bolded.

\subsection{Results}

\begin{table}[H]
\centering
\caption{Performance of the Progressive Multi-Modal Framework Across Different Datasets and Tasks}
\label{tab:threshold_results}
\resizebox{\linewidth}{!}{
\begin{tabular}{lllllllll}
\toprule
Dataset & Task & Stage & Modalities & Ratio (\%) & Accuracy & F1 & AUC & Threshold \\
\midrule

\multirow{9}{*}{ADNI}
& \multirow{3}{*}{CN vs AD}
& 1 & CLI & 50.00 & \multirow{3}{*}{0.9068} & \multirow{3}{*}{0.8809} & \multirow{3}{*}{0.8645} & 0.495 \\
& & 2 & CLI + MRI & 26.27 & & & & 0.015 \\
& & 3 & CLI + MRI + PET & 23.73 & & & & -- \\
\cline{2-9}

& \multirow{3}{*}{CN vs MCI}
& 1 & CLI & 64.52 & \multirow{3}{*}{0.8097} & \multirow{3}{*}{0.8025} & \multirow{3}{*}{0.8079} & 0.411 \\
& & 2 & CLI + MRI & 33.55 & & & & 0.436 \\
& & 3 & CLI + MRI + PET & 1.94 & & & & -- \\
\cline{2-9}

& \multirow{3}{*}{MCI vs AD}
& 1 & CLI & 58.51 & \multirow{3}{*}{0.7234} & \multirow{3}{*}{0.6727} & \multirow{3}{*}{0.6697} & 0.347 \\
& & 2 & CLI + MRI & 12.77 & & & & 0.347 \\
& & 3 & CLI + MRI + PET & 28.72 & & & & -- \\
\midrule

\multirow{9}{*}{AIBL}
& \multirow{3}{*}{CN vs AD}
& 1 & CLI & 54.50 & \multirow{3}{*}{0.9450} & \multirow{3}{*}{0.7955} & \multirow{3}{*}{0.7445} & 0.198 \\
& & 2 & CLI + MRI & 21.50 & & & & 0.149 \\
& & 3 & CLI + MRI + PET & 24.00 & & & & -- \\
\cline{2-9}

& \multirow{3}{*}{CN vs MCI}
& 1 & CLI & 84.44 & \multirow{3}{*}{0.8889} & \multirow{3}{*}{0.5854} & \multirow{3}{*}{0.5651} & 0.045 \\
& & 2 & CLI + MRI & 5.56 & & & & 0.084 \\
& & 3 & CLI + MRI + PET & 10.00 & & & & -- \\
\cline{2-9}

& \multirow{3}{*}{MCI vs AD}
& 1 & CLI & 4.26 & \multirow{3}{*}{0.6809} & \multirow{3}{*}{0.6713} & \multirow{3}{*}{0.6704} & 0.416 \\
& & 2 & CLI + MRI & 95.74 & & & & 0.451 \\
& & 3 & CLI + MRI + PET & 0.00 & & & & -- \\
\midrule

\multirow{9}{*}{OASIS}
& \multirow{3}{*}{CN vs AD}
& 1 & CLI & 88.04 & \multirow{3}{*}{0.9336} & \multirow{3}{*}{0.7548} & \multirow{3}{*}{0.7131} & 0.045 \\
& & 2 & CLI + MRI & 5.32 & & & & 0.079 \\
& & 3 & CLI + MRI + PET & 6.64 & & & & -- \\
\cline{2-9}

& \multirow{3}{*}{CN vs MCI}
& 1 & CLI & 100.00 & \multirow{3}{*}{0.9900} & \multirow{3}{*}{0.4975} & \multirow{3}{*}{0.5000} & 0.104 \\
& & 2 & CLI + MRI & 0.00 & & & & 0.010 \\
& & 3 & CLI + MRI + PET & 0.00 & & & & -- \\
\cline{2-9}

& \multirow{3}{*}{MCI vs AD}
& 1 & CLI & 58.33 & \multirow{3}{*}{0.7500} & \multirow{3}{*}{0.6783} & \multirow{3}{*}{0.7833} & 0.257 \\
& & 2 & CLI + MRI & 8.33 & & & & 0.421 \\
& & 3 & CLI + MRI + PET & 33.33 & & & & -- \\
\bottomrule
\end{tabular}
}
\end{table}

The accuracy, F1, and AUC columns are reused in Table \ref{tab:all_tasks_results_adni},\ref{tab:all_tasks_results_AIBL},\ref{tab:all_tasks_results_oasis} to perform more detailed analysis. From the experimental performance on three different datasets in Table \ref{tab:threshold_results}, at least 50\% of the dataset is qualified for an accurate decision, except for MCI vs AD performing in AIBL and at least 65\% of PET data are saved. To provide a better illustration of how many resources are saved in Table \ref{tab:threshold_results}, we will do a cost analysis on the real results. For this analysis, we will use \$3000 for the PET image \cite{mitka2013pet17} and \$558 for the MRI image. The total price of the PET image and the MRI image is \$3558. We computed the average imaging cost saving of at least 100 samples using the Qualified Samples (\%) as weighting. For example, with the task CN vs AD in ADNI, 50\% qualified (uncertainty value below the assigned threshold value) when just using clinical data, and 26.27\% qualified when using the combination of clinical and MRI data. As a result, $50\% * \$3558$ as 50\% doesn't need to get the MRI and PET, and $26.27\%\times\$3000$ as 26.27\% doesn't need to get the PET. In total, \$2567.1 is saved on average for one person. The remaining calculation of total saving cost across all datasets with three different binary tasks is shown in Table \ref{tab:cost_tasks}. From each task, the minimum amount of savings is \$2567.1, \$3171.1752, and \$2325.2814 for CN vs AD, CN vs MCI, and MCI vs AD, respectively. The minimum amount of savings across all tasks and datasets is \$2325.2814. From the statistics, we can see that on average, each person can save at least \$2300 to make a good AD diagnosis, which is an enormous saving for any person.

\begin{table}[H]
\centering
\caption{Total Saving Cost Across Datasets and Tasks}
\label{tab:cost_tasks}
\begin{tabular}{lccc}
\hline
Dataset & CN vs AD & CN vs MCI & MCI vs AD \\
\hline
ADNI  & \$2567.1000 & \$3302.1216 & \$2464.8858 \\
AIBL  & \$2584.1100 & \$3171.1752 & \$3023.7708 \\
OASIS & \$3292.0632 & \$3558.0000 & \$2325.2814 \\
\hline
\end{tabular}%
\end{table}

\begin{table}[H]
\centering
\caption{Comprehensive Performance Comparison Across Datasets, Models, and Tasks}
\label{tab:all_tasks_results_adni}
\resizebox{\textwidth}{!}{
\begin{tabular}{llccc ccc ccc}
\toprule

& & \multicolumn{3}{c}{CN vs AD} 
& \multicolumn{3}{c}{MCI vs AD} 
& \multicolumn{3}{c}{CN vs MCI} \\
\cmidrule(lr){3-5} \cmidrule(lr){6-8} \cmidrule(lr){9-11}

Dataset & Model 
& Acc & F1 & AUC 
& Acc & F1 & AUC 
& Acc & F1 & AUC \\
\midrule

\multirow{17}{*}{ADNI}
& KNN \cite{fix1950discriminatory34-2}  & 0.817 & 0.740 & 0.775 & 0.767 & 0.659 & 0.776 & 0.653 & 0.649 & 0.715 \\
& SVM \cite{vapnik199535-2}  & 0.902 & 0.869 & 0.962 & 0.850 & 0.811 & 0.840 & 0.659 & 0.655 & 0.751 \\
& LR \cite{tibshirani199636-2}   & 0.906 & 0.891 & \textbf{0.969} & \textbf{0.854} & \textbf{0.822} & \textbf{0.914} & 0.662 & 0.660 & 0.738 \\
& RF \cite{ho199537-2}   & 0.851 & 0.788 & 0.915 & 0.845 & 0.794 & 0.858 & 0.704 & 0.703 & 0.823 \\
& NN \cite{mcculloch194338-2}    & 0.902 & 0.869 & 0.906 & 0.840 & 0.797 & 0.809 & 0.666 & 0.660 & 0.740 \\
& GRridge \cite{van201639-2}  & 0.898 & 0.863 & 0.944 & 0.801 & 0.753 & 0.859 & 0.624 & 0.620 & 0.695 \\
& BPLSDA \cite{singh201940-2} & 0.894 & 0.868 & 0.965 & 0.830 & 0.767 & 0.853 & 0.666 & 0.648 & 0.740 \\
& BSPLSDA \cite{singh201940-2} & 0.830 & 0.798 & 0.879 & 0.830 & 0.767 & 0.853 & 0.720 & 0.716 & 0.802 \\
& MOGONET \cite{mogonet202122-2}  & 0.906 & 0.886 & 0.961 & 0.845 & 0.799 & 0.850 & \textbf{0.820} & \textbf{0.818} & \textbf{0.862} \\
& TMC \cite{han2022trusted10-2}  & 0.881 & 0.855 & 0.929 & 0.840 & 0.801 & 0.827 & 0.711 & 0.709 & 0.814 \\
& CF \cite{hong202041-2}    & 0.894 & 0.864 & 0.950 & 0.782 & 0.746 & 0.814 & 0.781 & 0.779 & 0.849 \\
& GMU \cite{arevalo201742-2}  & \textbf{0.915} & \textbf{0.892} & 0.949 & 0.835 & 0.804 & 0.826 & 0.662 & 0.660 & 0.762 \\
& MMDynamic \cite{han202243-2}  & 0.902 & 0.880 & 0.951 & 0.757 & 0.726 & 0.826 & 0.730 & 0.726 & 0.818 \\
& MLCLNet \cite{zheng202344-2}  & 0.911 & 0.890 & 0.940 & 0.820 & 0.789 & 0.844 & 0.704 & 0.697 & 0.795 \\
& DCP \cite{liu202145-2}  & 0.902 & 0.876 & 0.947 & 0.796 & 0.773 & 0.847 & 0.749 & 0.748 & 0.828 \\
& CLCLSA \cite{clclsa202423-2}   & 0.885 & 0.854 & 0.947 & 0.835 & 0.802 & 0.872 & 0.727 & 0.718 & 0.822 \\
& \textbf{ProMUSE} & 0.907 & 0.881 & 0.865 & 0.723 & 0.673 & 0.670 & 0.810 & 0.803 & 0.808 \\

\bottomrule
\end{tabular}
}
\end{table}

\begin{table}[H]
\centering
\caption{Comprehensive Performance Comparison Across Datasets, Models, and Tasks}
\label{tab:all_tasks_results_AIBL}
\resizebox{\textwidth}{!}{
\begin{tabular}{llccc ccc ccc}
\toprule

& & \multicolumn{3}{c}{CN vs AD} 
& \multicolumn{3}{c}{MCI vs AD} 
& \multicolumn{3}{c}{CN vs MCI} \\
\cmidrule(lr){3-5} \cmidrule(lr){6-8} \cmidrule(lr){9-11}

Dataset & Model 
& Acc & F1 & AUC 
& Acc & F1 & AUC 
& Acc & F1 & AUC \\
\midrule

\multirow{17}{*}{AIBL}
& KNN \cite{fix1950discriminatory34-2}  & 0.897 & 0.639 & 0.891 & 0.660 & 0.659 & 0.754 & 0.798 & 0.470 & 0.523 \\
& SVM \cite{vapnik199535-2}  & 0.938 & 0.825 & \textbf{0.989}  & 0.736 & 0.736 & 0.808 & 0.871 & 0.505 & 0.769 \\
& LR \cite{tibshirani199636-2}   & 0.959 & 0.908 & \textbf{0.989}  & 0.736 & 0.736 & 0.794 & 0.787 & 0.647 & 0.775 \\
& RF \cite{ho199537-2}   & \textbf{0.964} & 0.909 & \textbf{0.989}  & 0.660 & 0.659 & 0.779 & 0.871 & 0.465 & 0.752 \\
& NN \cite{mcculloch194338-2}    & 0.933 & 0.830 & 0.935 & 0.736 & 0.733 & 0.796 & 0.860 & 0.558 & 0.593 \\
& GRridge \cite{van201639-2}  & 0.944 & 0.850 & 0.920 & 0.755 & 0.753 & 0.811 & 0.831 & 0.611 & 0.725 \\
& BPLSDA \cite{singh201940-2} & 0.954 & 0.877 & 0.977 & 0.717 & 0.717 & 0.841 & 0.882 & 0.629 & 0.756 \\
& BSPLSDA \cite{singh201940-2} & 0.959 & 0.898 & 0.979 & 0.698 & 0.685 & 0.789 & 0.871 & 0.594 & 0.753 \\
& MOGONET \cite{mogonet202122-2}  & 0.944 & 0.872 & 0.984 & \textbf{0.792}  & \textbf{0.792} & 0.846 & 0.843 & 0.606 & \textbf{0.798} \\
& TMC \cite{han2022trusted10-2}  & 0.949 & 0.881 & 0.977 & 0.736 & 0.736 & 0.840 & 0.876 & \textbf{0.703}  & 0.797 \\
& CF \cite{hong202041-2}    & 0.959 & \textbf{0.911}  & 0.986 & 0.755 & 0.753 & 0.799 & 0.848 & 0.657 & 0.783 \\
& GMU \cite{arevalo201742-2}  & 0.944 & 0.872 & 0.980 & 0.736 & 0.736 & 0.800 & 0.837 & 0.645 & 0.731 \\
& MMDynamic \cite{han202243-2}  & 0.944 & 0.867 & 0.966 & 0.717 & 0.717 & 0.761 & 0.809 & 0.650 & 0.785 \\
& MLCLNet \cite{zheng202344-2}  & 0.954 & 0.895 & 0.968 & 0.679 & 0.675 & 0.771 & 0.860 & 0.682 & 0.777 \\
& DCP \cite{liu202145-2}  & 0.954 & 0.891 & 0.985 & 0.736 & 0.731 & \textbf{0.860}  & 0.843 & 0.606 & 0.738 \\
& CLCLSA \cite{clclsa202423-2}   & 0.938 & 0.852 & 0.974 & 0.755 & 0.754 & 0.813 & 0.865 & 0.689 & 0.744 \\
& \textbf{ProMUSE} & 0.945 & 0.796 & 0.745 & 0.681 & 0.671 & 0.670 & \textbf{0.889}  & 0.585 & 0.565 \\

\bottomrule
\end{tabular}
}
\end{table}

\begin{table}[H]
\centering
\caption{Comprehensive Performance Comparison Across Datasets, Models, and Tasks}
\label{tab:all_tasks_results_oasis}
\resizebox{\textwidth}{!}{
\begin{tabular}{llccc ccc ccc}
\toprule

& & \multicolumn{3}{c}{CN vs AD} 
& \multicolumn{3}{c}{MCI vs AD} 
& \multicolumn{3}{c}{CN vs MCI} \\
\cmidrule(lr){3-5} \cmidrule(lr){6-8} \cmidrule(lr){9-11}

Dataset & Model 
& Acc & F1 & AUC 
& Acc & F1 & AUC 
& Acc & F1 & AUC \\
\midrule

\multirow{17}{*}{OASIS}
& KNN \cite{fix1950discriminatory34-2}  & 0.902 & 0.599 & 0.676 & \textbf{0.838} & 0.456 & 0.449 & 0.983 & 0.496 & 0.608 \\
& SVM \cite{vapnik199535-2}  & 0.915 & 0.727 & 0.753 & 0.757 & 0.431 & 0.527 & 0.983 & 0.496 & 0.715 \\
& LR \cite{tibshirani199636-2}   & 0.892 & 0.700 & 0.798 & 0.784 & 0.538 & 0.565 & 0.970 & 0.492 & 0.749 \\
& RF \cite{ho199537-2}   & 0.899 & 0.643 & 0.883 & 0.838 & 0.456 & 0.683 & 0.983 & 0.496 & 0.702 \\
& NN \cite{mcculloch194338-2}    & 0.918 & 0.700 & 0.784 & 0.730 & 0.560 & 0.672 & 0.977 & 0.494 & 0.463 \\
& GRridge \cite{van201639-2}  & 0.902 & 0.705 & 0.698 & 0.757 & 0.431 & 0.516 & 0.980 & 0.495 & 0.491 \\
& BPLSDA \cite{singh201940-2} & 0.918 & 0.733 & 0.790 & 0.730 & 0.503 & 0.540 & 0.983 & 0.496 & 0.780 \\
& BSPLSDA \cite{singh201940-2} & 0.912 & 0.662 & \textbf{0.893} & 0.730 & 0.503 & 0.540 & 0.983 & 0.496 & 0.780 \\
& MOGONET \cite{mogonet202122-2}  & 0.915 & 0.752 & 0.885 & 0.514 & 0.495 & 0.634 & 0.973 & 0.493 & 0.545 \\
& TMC \cite{han2022trusted10-2}  & 0.922 & 0.729 & 0.866 & 0.649 & 0.393 & 0.629 & 0.977 & 0.494 & 0.659 \\
& CF \cite{hong202041-2}    & 0.908 & 0.695 & 0.880 & 0.514 & 0.495 & 0.683 & 0.980 & 0.495 & 0.632 \\
& GMU \cite{arevalo201742-2}  & 0.899 & 0.726 & 0.792 & 0.595 & 0.427 & 0.489 & 0.980 & 0.495 & 0.520 \\
& MMDynamic \cite{han202243-2}  & 0.902 & 0.723 & 0.842 & 0.676 & 0.471 & 0.586 & 0.906 & 0.509 & 0.657 \\
& MLCLNet \cite{zheng202344-2}  & 0.912 & 0.730 & 0.837 & 0.432 & 0.417 & 0.575 & 0.980 & 0.495 & \textbf{0.838}  \\
& DCP \cite{liu202145-2}  & 0.908 & 0.725 & 0.880 & 0.676 & 0.403 & 0.688 & 0.970 & 0.492 & 0.685 \\
& CLCLSA \cite{clclsa202423-2}   & 0.895 & 0.713 & 0.856 & 0.784 & 0.439 & 0.473 & 0.883 & \textbf{0.542} & 0.640 \\
& \textbf{ProMUSE} & \textbf{0.934}  & \textbf{0.755} & 0.713 & 0.750 & \textbf{0.678}  & \textbf{0.783}  & \textbf{0.990}  & 0.498 & 0.5 \\

\bottomrule
\end{tabular}
}
\end{table}

In table \ref{tab:all_tasks_results_adni},\ref{tab:all_tasks_results_AIBL},\ref{tab:all_tasks_results_oasis}, ProMUSE performed very well in measuring accuracy in CN vs AD and CN vs MCI tasks, with the difference in the highest accuracy between 0 and 0.019. In ADNI, ProMUSE receives 0.907 accuracy compared to the highest accuracy of 0.915 OF GMU in CN vs AD, with a competitive 0.881 F1 and 0.865 AUC. With CN vs MCI in the same dataset, ProMUSE has 0.810 accuracy compared to the highest accuracy of 0.820 OF MOGONET with a high 0.803 F1 and 0.808 AUC. In AIBL, ProMUSE accuracy is very high at 0.945  (less than 0.019 of the highest accuracy, 0.964 of RF) in CN vs AD and 0.889, the highest, in CN vs MCI. However, its F1 and AUC scores are lower compared to the most competitive ones. In OASIS, the accuracies of ProMUSE are the highest in both CN vs AD with 0.934 and CN vs MCI with 0.990, followed by competitive F1 0.755 in CN vs AD and F1 0.498 in CN vs MCI. However, the AUC is low compared to the highest ones from BSPLSDA in CN vs AD and MLCLNET in CN vs MCI. In MCI vs AD, the results are not positive in ADNI and AIBL. IN ADNI, ProMUSE has 0.723 accuracy, 0.673 F1, and 0.670 AUC compared to the highest accuracy of 0.854, the highest F1 of 0.822, and the highest AUC of 0.914 from LR. IN AIBL, ProMUSE has 0.681 accuracy, 0.671 F1, and 0.670 AUC compared to the highest accuracy of 0.792 from MOGONET, the highest F1 of 0.792 from MOGONET, and the highest AUC of 0.860 from DCP. Nevertheless, with MCI vs AD in OASIS, ProMUSE shows a positive result with the highest F1 of 0.678 and the highest AUC of 0.783, and a good accuracy of 0.750 compared to the highest of 0.838 from KNN. Overall, ProMuse shows an extraordinary performance across all 3 datasets in CN vs AD and CN vs MCI, except MCI vs AD. 
\begin{figure}[H]
    \centering
    \includegraphics[width=1\textwidth]{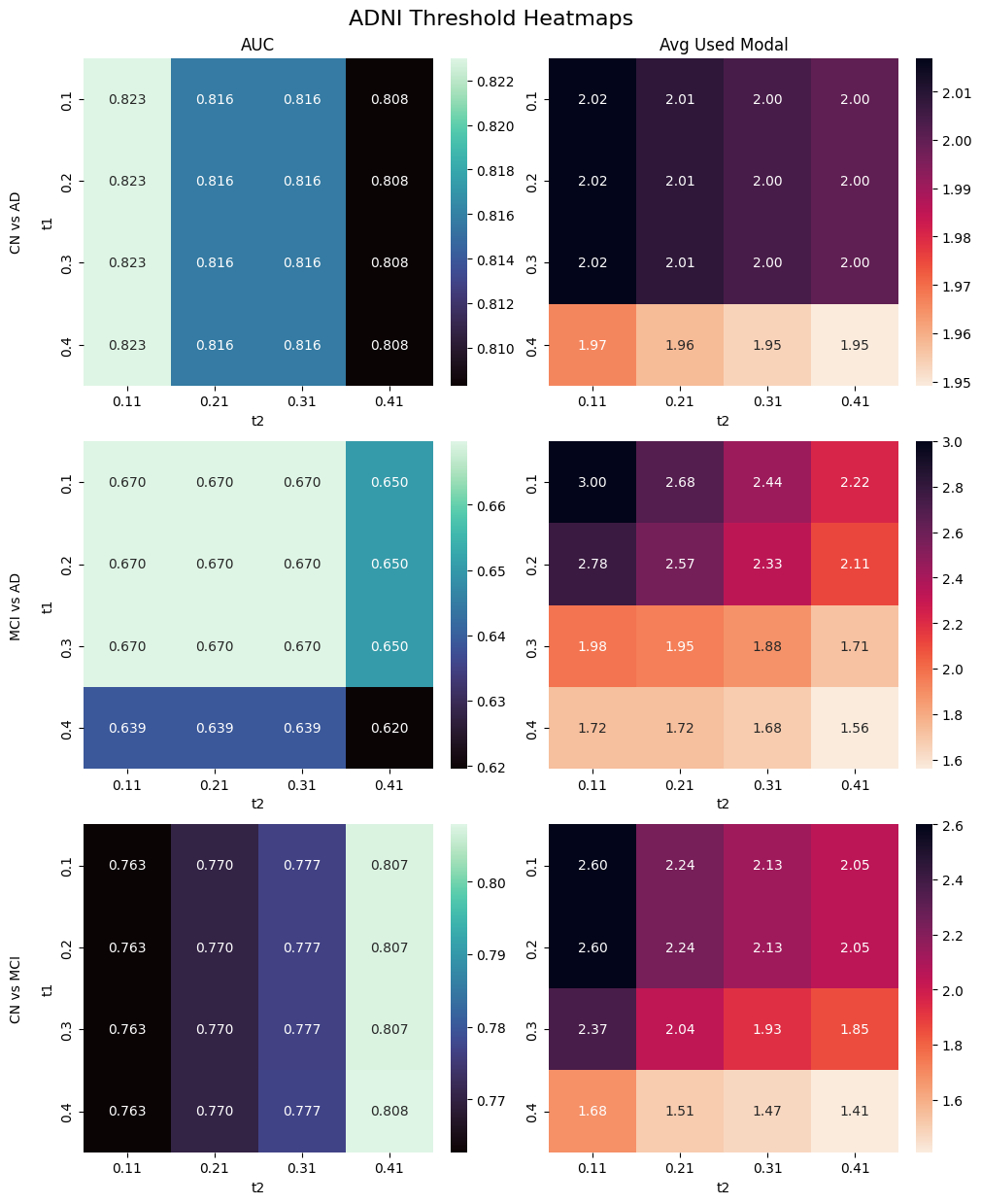}
    \caption{Diagram for the correlation between AUC and the average used modalities in the ADNI dataset}
    \label{fig:adni}
\end{figure}

\begin{figure}[H]
    \centering
    \includegraphics[width=1\textwidth]{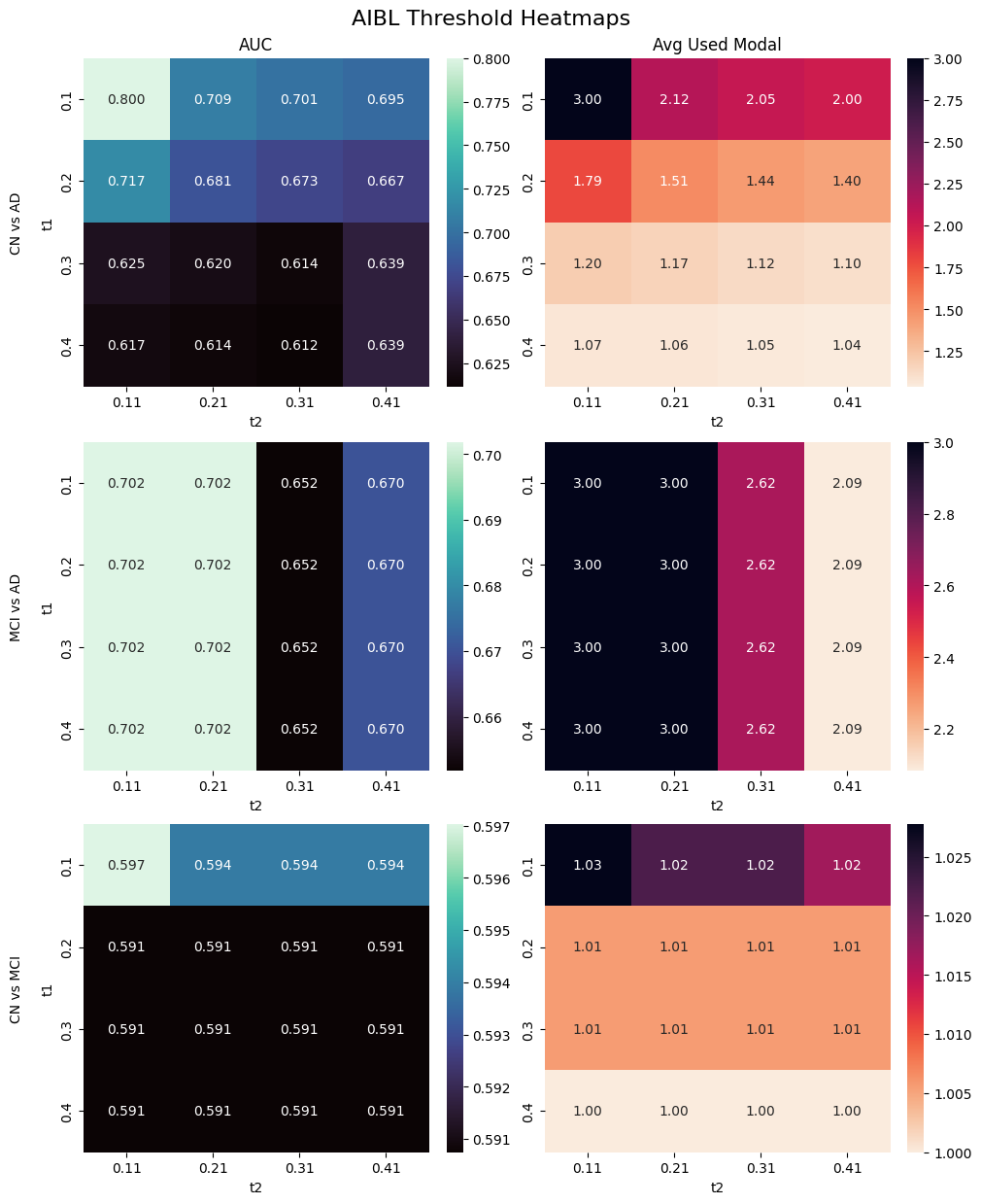}
    \caption{Diagram for the correlation between AUC and the average used modalities in the AIBL dataset}
    \label{fig:aibl}
\end{figure}

\begin{figure}[H]
    \centering
    \includegraphics[width=1\textwidth]{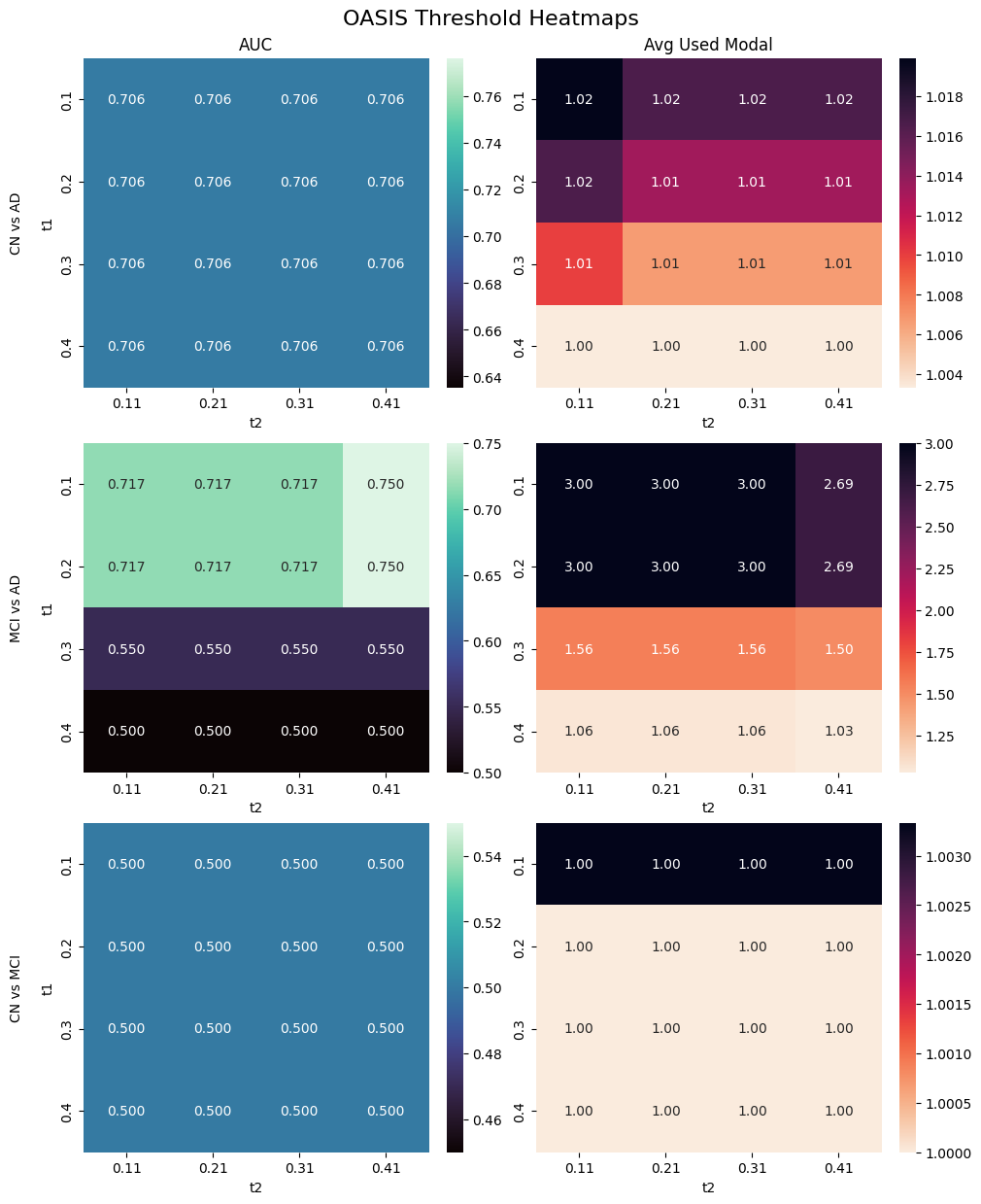}
    \caption{Diagram for the correlation between AUC and the average used modalities in the OASIS dataset}
    \label{fig:oasis}
\end{figure}

Fig \ref{fig:adni}, \ref{fig:aibl}, and \ref{fig:oasis} show the relationship between AUC and the average used modalities in prediction in ADNI, AIBL, and OASIS. t1(\(\tau_1\)) is the first assigned threshold and t2(\(\tau_2\)) is the second assigned threshold. AUC is calculated using the framework in Fig. \ref{fig:uncertainty_framework}. Overall, with lower t1 and t2, the average used modalities is higher. For example, with MCI vs AD in ADNI and the same t1 at 0.1, t2 at 0.11,0.21,0.31,0.41, used 3, 2.68, 2.44, 2.22 modalities on average. This trend applies across all 3 different datasets and all tasks. As analyzed in table \ref{tab:all_tasks_results_adni},\ref{tab:all_tasks_results_AIBL},\ref{tab:all_tasks_results_oasis}, ProMUSE performed poorly with the MCI vs AD task, which is illustrated very obviously in the number of average modalities used in these figures. With low t1 0.1 and 0.2, all three datasets need to use at least 2 to 3 modalities as the uncertainty in the decision is high. From an overall perspective, more modalities used means higher accuracy. However, there are some cases where more modalities results in lower accuracies. For example, CN vs MCI in ADNI, at t1 0.1 and t2 0.41, 0.807 AUC is achieved with 2.05 average used modalites, while at t1 0.1 and t2 0.11, 0.763 AUC is achieved with 2.60 average used modalites. This shows the fact that lower uncertainty doesn't always result in higher overall accuracy. To achieve a high accuracy, an optimal uncertainty is needed.

Despite the promising results, our experiment is not validated on real-world applications, so further testing must be done to apply this framework in normal life. In the future, more algorithms can be applied to make the uncertainty degree more robust.

\section{CONCLUSION}
We propose ProMUSE, a staged cost-efficient uncertainty-aware multi-modal framework for AD prediction. This framework uses resources progressively and still maintains high accuracy by utilizing the uncertainty at single-modality levels. For multiple modalities, we apply the Dempster-Shafer theory to combine the beliefs and uncertainties of the modalities effectively. 

An additional modality is only added when there is insufficient evidence. This helps save a lot of resources for people performing AD prediction. The superior performance through 3 different datasets shows that, on average, thousands of dollars can be saved. All of these results are proof that the framework is working effectively on real-world data. Although positive results are shown, the need to validate them in a real-world application on real humans remains.

\subsection*{Author Contributions} 
L. Doan, B. Chen, and E. Litton: conceptualization, methodology, software, and manuscript writing. H. Huang: manuscript review and editing. J. Huang and Y. Xie: manuscript review and editing. N. Narayanan: clinical expertise, interpretation of results, and manuscript review and editing. W. Zhou: conceptualization and manuscript writing. C. Zhao: conceptualization, funding acquisition, project administration, supervision, and manuscript writing.

\subsection*{Funding}
This work was partially supported by the American Heart Association under Award No. 25AIREA1377168 (PI: Chen Zhao)

\subsection*{Conflicts of Interest}
The authors declare that there is no conflict of interest regarding the publication of this article

\subsection*{Data Availability}
The raw data belong to ADNI, AIBL and OASIS3 data providers.

\printbibliography

@article{dukart2013relationship6,
  author = {Dukart, J. and Mueller, K. and Villringer, A. and Kherif, F. and Draganski, B. and Frackowiak, R. and Schroeter, M. L.},
  title = {Relationship between imaging biomarkers, age, progression and symptom severity in Alzheimer’s disease},
  journal = {NeuroImage: Clinical},
  volume = {3},
  pages = {84--94},
  year = {2013},
  pmid = {24179852}
}

@article{song2021effective9,
  author = {Song, J. and Zheng, J. and Li, P. and Lu, X. and Zhu, G. and Shen, P.},
  title = {An Effective Multimodal Image Fusion Method Using MRI and PET for Alzheimer’s Disease Diagnosis},
  journal = {Frontiers in Digital Health},
  volume = {3},
  pages = {637386},
  year = {2021},
  pmid = {34713109}
}

@inproceedings{huang2021radiogenomics11,
  author = {Huang, Y. and Li, L. and Jiang, J.},
  title = {Radiogenomics of Alzheimer’s disease: exploring gene related metabolic imaging markers},
  booktitle = {2021 43rd Annual International Conference of the IEEE Engineering in Medicine \& Biology Society (EMBC)},
  pages = {5772--5775},
  year = {2021},
  organization = {IEEE}
}

@article{matthews2019racial16,
  author = {Matthews, K. A. and Xu, W. and Gaglioti, A. H. and Holt, J. B. and Croft, J. B. and Mack, D. and McGuire, L. C.},
  title = {Racial and ethnic estimates of Alzheimer’s disease and related dementias in the United States (2015--2060) in adults aged $\geq$65 years},
  journal = {Alzheimer's \& Dementia},
  volume = {15},
  number = {1},
  pages = {17--24},
  year = {2019},
  pmid = {30243772}
}

@article{mitka2013pet17,
  author = {Mitka, M.},
  title = {PET Imaging for Alzheimer Disease: Are Its Benefits Worth the Cost?},
  journal = {JAMA},
  volume = {309},
  number = {11},
  pages = {1099},
  year = {2013},
  pmid = {23512036}
}

@article{mahaman2022biomarkers18,
  author = {Mahaman, Y. A. R. and Embaye, K. S. and Huang, F. and Li, L. and Zhu, F. and Wang, J.-Z. and Liu, R. and Feng, J. and Wang, X.},
  title = {Biomarkers used in Alzheimer’s disease diagnosis, treatment, and prevention},
  journal = {Ageing Research Reviews},
  volume = {74},
  pages = {101544},
  year = {2022},
  pmid = {34933129}
}

@article{veitch2019understanding19,
  author = {Veitch, D. P. and Weiner, M. W. and Aisen, P. S. and Beckett, L. A. and Cairns, N. J. and Green, R. C. and Harvey, D. and Jack, C. R. and Jagust, W. and Morris, J. C. and Petersen, R. C. and Saykin, A. J. and Shaw, L. M. and Toga, A. W. and Trojanowski, J. Q.},
  title = {Understanding disease progression and improving Alzheimer’s disease clinical trials: Recent highlights from the Alzheimer’s Disease Neuroimaging Initiative},
  journal = {Alzheimer’s \& Dementia},
  volume = {15},
  number = {1},
  pages = {106--152},
  year = {2019},
  pmid = {30321505}
}

@article{chandra2019mri23,
  author = {Chandra, A. and Dervenoulas, G. and Politis, M.},
  title = {Magnetic resonance imaging in Alzheimer’s disease and mild cognitive impairment},
  journal = {Journal of Neurology},
  volume = {266},
  number = {6},
  pages = {1293--1302},
  year = {2019},
  pmid = {30120563}
}

@article{minoshima1997metabolic24,
  author = {Minoshima, S. and Giordani, B. and Berent, S. and Frey, K. A. and Foster, N. L. and Kuhl, D. E.},
  title = {Metabolic reduction in the posterior cingulate cortex in very early Alzheimer’s disease},
  journal = {Annals of Neurology},
  volume = {42},
  number = {1},
  pages = {85--94},
  year = {1997},
  pmid = {9225689}
}

@article{yang2022combining25,
  author = {Yang, F. and Jiang, J. and Alberts, I. and Wang, M. and Li, T. and Sun, X. and Rominger, A. and Zuo, C. and Shi, K.},
  title = {Combining PET with MRI to improve predictions of progression from mild cognitive impairment to Alzheimer’s disease: an exploratory radiomic analysis study},
  journal = {Annals of Translational Medicine},
  volume = {10},
  number = {9},
  pages = {513--513},
  year = {2022},
  pmid = {35928737}
}

@article{castellano2024automated26,
  author = {Castellano, G. and Esposito, A. and Lella, E. and Montanaro, G. and Vessio, G.},
  title = {Automated detection of Alzheimer’s disease: a multi-modal approach with 3D MRI and amyloid PET},
  journal = {Scientific Reports},
  volume = {14},
  number = {1},
  pages = {5210},
  year = {2024},
  pmid = {38433282}
}

@article{venugopalan2021multimodal29,
  author = {Venugopalan, J. and Tong, L. and Hassanzadeh, H. R. and Wang, M. D.},
  title = {Multimodal deep learning models for early detection of Alzheimer’s disease stage},
  journal = {Scientific Reports},
  volume = {11},
  number = {1},
  pages = {1--13},
  year = {2021},
  pmid = {33547343}
}

@article{zhang2019multimodal31,
  author = {Zhang, F. and Li, Z. and Zhang, B. and Du, H. and Wang, B. and Zhang, X.},
  title = {Multi-modal deep learning model for auxiliary diagnosis of Alzheimer’s disease},
  journal = {Neurocomputing},
  volume = {361},
  pages = {185--195},
  year = {2019}
}

@article{geldmacher2013prediagnosis32,
  author = {Geldmacher, D. S. and Kirson, N. Y. and Birnbaum, H. G. and Eapen, S. and Kantor, E. and Cummings, A. K. and Joish, V. N.},
  title = {Pre-Diagnosis Excess Acute Care Costs in Alzheimer’s Patients among a US Medicaid Population},
  journal = {Applied Health Economics and Health Policy},
  volume = {11},
  number = {4},
  pages = {407--413},
  year = {2013},
  pmid = {23700254}
}

@article{han2022trusted10-2,
  author = {Han, Z. and Zhang, C. and Fu, H. and Zhou, J.T.},
  title = {Trusted multi-view classification with dynamic evidential fusion},
  journal = {IEEE Trans. Pattern Anal. Mach. Intell.},
  volume = {45},
  year = {2022},
  pages = {2551--2566}
}

@book{josang201618-2,
  author = {J{\o}sang, A.},
  title = {Subjective Logic},
  publisher = {Springer},
  year = {2016}
}

@article{mogonet202122-2,
  author = {Wang, T. and Shao, W. and Huang, Z. and Tang, H. and Zhang, J. and Ding, Z. and Huang, K.},
  title = {MOGONET integrates multi-omics data using graph convolutional networks allowing patient classification and biomarker identification},
  journal = {Nature Communications},
  volume = {12},
  year = {2021},
  pages = {3445}
}

@article{clclsa202423-2,
  author = {Zhao, C. and Liu, A. and Zhang, X. and Cao, X. and Ding, Z. and Sha, Q. and Shen, H. and Deng, H.-W. and Zhou, W.},
  title = {CLCLSA: cross-omics linked embedding with contrastive learning and self-attention for integration with incomplete multi-omics data},
  journal = {Computers in Biology and Medicine},
  volume = {170},
  year = {2024},
  pages = {108058}
}

@book{cover199929-2,
  author = {Cover, T.M.},
  title = {Elements of Information Theory},
  publisher = {John Wiley \& Sons},
  year = {1999}
}

@article{vapnik199535-2,
  author = {Vapnik, V.},
  title = {Support-vector networks},
  journal = {Machine Learning},
  volume = {20},
  year = {1995},
  pages = {273--297}
}

@article{tibshirani199636-2,
  author = {Tibshirani, R.},
  title = {Regression shrinkage and selection via the lasso},
  journal = {JRSS B},
  volume = {58},
  year = {1996},
  pages = {267--288}
}

@inproceedings{ho199537-2,
  author = {Ho, T.K.},
  title = {Random decision forests},
  booktitle = {ICDAR},
  year = {1995},
  pages = {278--282}
}

@article{mcculloch194338-2,
  author = {McCulloch, W.S. and Pitts, W.},
  title = {A logical calculus of nervous activity},
  journal = {Bull. Math. Biophys.},
  volume = {5},
  year = {1943},
  pages = {115--133}
}

@article{van201639-2,
  author = {Van De Wiel, M.A. and Lien, T.G. and Verlaat, W. and van Wieringen, W.N. and Wilting, S.M.},
  title = {Better prediction by use of co-data},
  journal = {Statistics in Medicine},
  volume = {35},
  year = {2016},
  pages = {368--381}
}

@article{singh201940-2,
  author = {Singh, A. and Shannon, C.P. and Gautier, B. and Rohart, F. and Vacher, M. and Tebbutt, S.J. and L{\^e} Cao, K.-A.},
  title = {DIABLO integrative multi-omics},
  journal = {Bioinformatics},
  volume = {35},
  year = {2019},
  pages = {3055--3062}
}

@article{hong202041-2,
  author = {Hong, D. and Gao, L. and Yokoya, N. and Yao, J. and Chanussot, J. and Du, Q. and Zhang, B.},
  title = {More diverse means better multimodal learning},
  journal = {IEEE TGRS},
  volume = {59},
  year = {2020},
  pages = {4340--4354}
}

@article{arevalo201742-2,
  author = {Arevalo, J. and Solorio, T. and Montes-y-Gomez, M. and Gonzalez, F.A.},
  title = {Gated multimodal units},
  journal = {arXiv},
  year = {2017}
}

@inproceedings{han202243-2,
  author = {Han, Z. and Yang, F. and Huang, J. and Zhang, C. and Yao, J.},
  title = {Multimodal dynamics fusion},
  booktitle = {CVPR},
  year = {2022},
  pages = {20707--20717}
}

@inproceedings{zheng202344-2,
  author = {Zheng, X. and Tang, C. and Wan, Z. and Hu, C. and Zhang, W.},
  title = {Multi-level confidence learning},
  booktitle = {AAAI},
  year = {2023},
  pages = {11381--11389}
}

@article{liu202145-2,
  author = {Liu, J. and Zhuang, B. and Zhuang, Z. and Guo, Y. and Huang, J. and Zhu, J. and Tan, M.},
  title = {Discrimination-aware network pruning},
  journal = {IEEE TPAMI},
  volume = {44},
  year = {2021},
  pages = {4035--4051}
}

@incollection{dempster200830-2,
  author = {Dempster, A.P.},
  title = {Upper and Lower Probabilities Induced by a Multivalued Mapping},
  booktitle = {Classic Works of the Dempster-Shafer Theory of Belief Functions},
  publisher = {Springer},
  year = {2008},
  pages = {57--72}
}

@misc{fix1950discriminatory34-2,
  author       = {E. Fix},
  title        = {Discriminatory Analysis: Nonparametric Discrimination, Consistency Properties},
  howpublished = {USAF School of Aviation Medicine},
  year         = {1950},
  note         = {Technical Report}
}

@article{fischl2012freesurfer,
  author  = {Bruce Fischl},
  title   = {FreeSurfer},
  journal = {NeuroImage},
  volume  = {62},
  number  = {2},
  pages   = {774--781},
  year    = {2012}
}

@article{zhao2021lung,
  author  = {Chao Zhao and others},
  title   = {Lung Segmentation and Automatic Detection of COVID-19 Using Radiomic Features from Chest CT Images},
  journal = {Pattern Recognition},
  volume   = {119},
  pages    = {108071},
  year     = {2021}
}

@inproceedings{hamilton2017inductivegraphsage,
  author    = {William L. Hamilton and Rex Ying and Jure Leskovec},
  title     = {Inductive Representation Learning on Large Graphs},
  booktitle = {Advances in Neural Information Processing Systems (NeurIPS)},
  volume     = {30},
  year       = {2017}
}

@inproceedings{teerapittayanon2016branchynet5-2,
author    = {Teerapittayanon, Surat and McDanel, Bradley and Kung, H. T.},
title     = {BranchyNet: Fast Inference via Early Exiting from Deep Neural Networks},
booktitle = {Proceedings of the 23rd International Conference on Pattern Recognition (ICPR)},
year      = {2016},
pages      = {2464--2469},
publisher = {IEEE}
}

@inproceedings{contardo2016sequential6-2,
author    = {Contardo, Guillaume and Denoyer, Ludovic and Arti{`e}res, Thierry},
title     = {Sequential Cost-Sensitive Feature Acquisition},
booktitle = {International Symposium on Intelligent Data Analysis},
year      = {2016},
pages     = {284--294},
publisher = {Springer}
}

@article{an2022reinforcement7-2,
author  = {An, Chao and Zhou, Qiang and Yang, Shuo},
title   = {A Reinforcement Learning Guided Adaptive Cost-Sensitive Feature Acquisition Method},
journal = {Applied Soft Computing},
volume  = {117},
pages   = {108437},
year    = {2022}
}

@inproceedings{wu2018blockdrop12-2,
author    = {Wu, Zuxuan and Nagarajan, Tushar and Kumar, Abhishek and Rennie, Steven and Davis, Larry S. and Grauman, Kristen and Feris, Rogerio},
title     = {BlockDrop: Dynamic Inference Paths in Residual Networks},
booktitle = {Proceedings of the IEEE Conference on Computer Vision and Pattern Recognition (CVPR)},
year      = {2018},
pages     = {8817--8826}
}

@article{zhou2024aidriven46,
  author    = {Zhou, Wei and Wang, Yang and Wu, Yu and Li, Xiang and Liu, Hui and Wang, Hui and Zhang, Zhen and Huang, Hong},
  title     = {AI-driven Multimodal Precision Diagnosis and Progression Prediction of Alzheimer's Disease: Data Fusion Mechanisms, Clinical Applications, and Research Trends (2017--2024)},
  journal   = {Digital Health},
  volume     = {12},
  pages      = {20552076251412649},
  year       = {2026},
  doi        = {10.1177/20552076251412649}
}

@article{ayigit2022dementia47,
  author    = {Yi\u{g}it, Altu\u{g} and Ba\c{s}tanlar, Yal{\i}n and I\c{s}{\i}k, Zerrin},
  title     = {Dementia Diagnosis by Ensemble Deep Neural Networks Using FDG-PET Scans},
  journal   = {Signal, Image and Video Processing},
  volume    = {16},
  number    = {8},
  pages     = {2203--2210},
  year      = {2022},
  doi       = {10.1007/s11760-022-02185-4}
}

@article{gowda2022multimodal48,
  author    = {Mohan Gowda, V. and Arakeri, Megha P. and Raghu Ram Prasad, Vasireddy},
  title     = {Multimodal Classification Technique for Fall Detection of Alzheimer's Patients by Integration of a Novel Piezoelectric Crystal Accelerometer and Aluminum Gyroscope with Vision Data},
  journal   = {Advances in Materials Science and Engineering},
  volume    = {2022},
  pages     = {9258620},
  year      = {2022},
  doi       = {10.1155/2022/9258620}
}

@article{li2022integratingEI,
    author = {Li, Yan Chak and Wang, Linhua and Law, Jeffrey N and Murali, T M and Pandey, Gaurav},
    title = {Integrating multimodal data through interpretable heterogeneous ensembles},
    journal = {Bioinformatics Advances},
    volume = {2},
    number = {1},
    pages = {vbac065},
    year = {2022},
    month = {01},
    issn = {2635-0041},
    doi = {10.1093/bioadv/vbac065},
    url = {https://doi.org/10.1093/bioadv/vbac065},
    eprint = {https://academic.oup.com/bioinformaticsadvances/article-pdf/2/1/vbac065/47086253/vbac065.pdf},
}

@article{petersen2010adni46,
  author    = {Petersen, Ronald C. and Aisen, Paul S. and Beckett, Laurel A. and Donohue, Michael C. and Gamst, Anthony C. and Harvey, Danielle J. and Jack, Clifford R. Jr. and Jagust, William J. and Shaw, Leslie M. and Toga, Arthur W. and Trojanowski, John Q. and Weiner, Michael W.},
  title     = {Alzheimer's Disease Neuroimaging Initiative (ADNI): Clinical Characterization},
  journal   = {Neurology},
  volume    = {74},
  number    = {3},
  pages     = {201--209},
  year      = {2010},
  doi       = {10.1212/WNL.0b013e3181cb3e25}
}

@article{ellis2009aiblDataset,
  author    = {Ellis, Kathryn A. and Bush, Ashley I. and Darby, David and De Fazio, David and Foster, Jennifer and Hudson, Peter and Lautenschlager, Nicola T. and Lenzo, Nick and Martins, Ralph N. and Maruff, Paul and Masters, Colin and Milner, Anne and Pike, Kathryn and Rowe, Christopher and Savage, Greg and Szoeke, Cassandra and Taddei, Kevin and Villemagne, Victor and Woodward, Michael and Ames, David and AIBL Research Group},
  title     = {The Australian Imaging, Biomarkers and Lifestyle (AIBL) Study of Aging: Methodology and Baseline Characteristics of 1112 Individuals Recruited for a Longitudinal Study of Alzheimer's Disease},
  journal   = {International Psychogeriatrics},
  volume    = {21},
  number    = {4},
  pages     = {672--687},
  year      = {2009},
  doi       = {10.1017/S1041610209009405}
}

@article{marcus2010oasisDataset,
  author    = {Marcus, Daniel S. and Fotenos, Alexander F. and Csernansky, John G. and Morris, John C. and Buckner, Randy L.},
  title     = {Open Access Series of Imaging Studies (OASIS): Longitudinal MRI Data in Nondemented and Demented Older Adults},
  journal   = {Journal of Cognitive Neuroscience},
  volume    = {22},
  number    = {12},
  pages     = {2677--2684},
  year      = {2010},
  doi       = {10.1162/jocn.2009.21407}
}

\end{document}